\pdfoutput=1

\documentclass[11pt]{article}

\usepackage{emnlp2021}

\usepackage{times}
\usepackage{latexsym}

\usepackage{breqn}
\usepackage{amsmath}
\usepackage{url}
\usepackage{subcaption}
\usepackage{makecell}

\usepackage[T1]{fontenc}

\usepackage[utf8]{inputenc}

\usepackage{microtype}
\usepackage{comment}

\usepackage{soul}
\usepackage{epstopdf}
\usepackage{booktabs}
\usepackage{graphicx}

\title{A Comparative Study on Automatic Coding of Medical Letters with Explainability}

\author{
         Jamie Glen $^\alpha$, Lifeng Han $^\beta$, Paul Rayson $^\alpha$,    
          \and Goran Nenadic $^\beta$\\
          $^\alpha$Lancaster University |  $^\beta$The University of Manchester, UK \\ 
         \\ {\tt jamie.a.glen@outlook.com, p.rayson@lancaster.ac.uk } 
         \\
         {\tt  
         lifeng.han, g.nenadic@manchester.ac.uk}         }

\begin{document}
\maketitle{}

\begin{abstract}
This study aims to explore the implementation of Natural Language Processing (NLP) and machine learning (ML) techniques to automate the coding of medical letters with visualised explainability and light-weighted local computer settings.
Currently in clinical settings, coding is a manual process that involves assigning codes to each condition, procedure, and medication in a patient's paperwork (e.g., 56265001 heart disease using SNOMED CT code). 
There are preliminary research on automatic coding in this field using state-of-the-art ML models; however, due to the complexity and size of the models, the real-world deployment is not achieved. 
To further facilitate the possibility of automatic coding practice, we explore some solutions in a local computer setting; in addition, we explore the function of explainability for transparency of AI models.
We used the publicly available MIMIC-III database and the HAN/HLAN network models for ICD code prediction purposes. We also experimented with the mapping between ICD and SNOMED CT knowledge bases.
In our experiments, the models
provided useful information for 97.98\% of codes. 
The result of this investigation can shed some light on implementing automatic clinical coding in practice, such as in hospital settings, on the local computers used by clinicians, project page \url{https://github.com/Glenj01/Medical-Coding}.


\end{abstract}

\section{Introduction}

The coding of medical letters is currently something that is completed manually in advanced healthcare systems such as the UK and the US \footnote{NHS UK \url{https://www.nhs.uk/}}. It involves professionals reviewing the paperwork for a patient's hospital visit or appointment and assigning specific codes to the conditions, diseases, procedures, and medications in the letters. This study aims to examine the potential automation of this process using Natural Language Processing (NLP) and Machine-Learning (ML) techniques, to create a prototype that could be used alongside the coders to speed up the coding process and to explore if such a system could be integrated into the real practice. 

Clinical codes are used to remove ambiguity in the language of the letters, provide easily generated statistics, give a standardised way to represent medical concepts and allow the NHS’s Electronic Health Record (EHR) system to process and store the codes more easily \cite{NHS-Digital2023}. Also, in the case of private healthcare providers, coding can make it easier to keep track of billing \footnote{\url{https://www.ashfordstpeters.nhs.uk/clinical-coding}}. 
To do this, the coder takes a medical letter as input, which can be anything from a prescription request to a hospital discharge summary, and outputs potential codes from a designated terminology and/or classification system. 
The NHS ‘fundamental information standard’ is the  ``Systemised Nomenclature of Medicine – Clinical Terms'' (\textit{aka} SNOMED-CT) terminology system, which uses ‘concepts’ to represent clinical thoughts. 
Each concept is paired with a ‘Concept Id’ – a unique numerical identifier e.g., 56265001 heart disease (disorder) - which is then arranged by relationships into hierarchies from the general to the more detailed \cite{NHS-Digital2023}.
It is worth noting that SNOMED is not the only system used for coding. 
The other system relevant to this work is the International Classification of Diseases (ICD), specifically ICD-9 \footnote{\url{https://www.cdc.gov/nchs/icd/icd9cm.htm}}. 
This was the official system used to code diagnoses and procedures in the US.
While SNOMED is a terminology system that has a comprehensive scope, covering every illness, event, symptom, procedure, test, organism, substance, and medicine, ICD is a classification system with a scope of just classifying diagnoses and procedures. 
In the NHS UK, coding is a significant issue because it takes time, energy, and resources away from an already underfunded and overworked system. There have been efforts to solve this by having dedicated clinical coding departments in larger hospitals \footnote{\url{https://www.stepintothenhs.nhs.uk/careers/clinical-coder}}; however, in most smaller practices, it is still the medical professionals who will do the coding. 
It takes the average coder 7-8 minutes to code each case, and a dedicated department of 25-30 coders usually codes more than 20,000 cases monthly \cite{dong2022automated_11}. 
Even so, there is almost always a backlog of cases to be coded, which has been known to extend over a year. 
It is estimated that AI applications in the healthcare industry have the potential to free up 1.944 million hours each year for healthcare professionals, with the biggest cut being taken from AI in virtual health assistance (such as automated medical coding) at 1.145 million hours \cite{13_Biundo2020socio}.
Clinical coding is such a challenging task due to two main concerns. The first is that the classification systems are complex and dynamic. The international edition of SNOMED contains 352,567 concepts \footnote{Five Step Briefing, SNOMED international \url{https://www.snomed.org/five-step-briefing}}, and while it should be noted that not all of these are diagnoses, finding the correct code can be challenging. 
The other issue is that there is no consistent structure in the documents to be coded. They can be notational, lengthy, and incomplete in addition to being full of abbreviations and symbols. 
Since all the coding is done manually, the human factor must also be considered. A study by \newcite{burns2012systematic_12} found that the median accuracy of coders under evaluation was 83.2\%. It should be noted that this was with an interquartile range of 67.3\% - 92.1\%, which further proves the issue of inconsistency with human coding. 

This paper explores the potential of replacing the time-consuming process of manually coding letters with a program that automatically assigns codes to letters in a local computer setup. 
In the following sections, this paper will explore the background of automated medical coding, explain the implementation choices and issues encountered with this investigation, review the testing methods and results, and conclude by discussing the implications of these findings and the potential future of medical coding. 

\section{Backgrounds and Related Work}

The background session will be presented in two sections. The first section, pre-neural networks, will focus on the early attempts at automated medical coding, how they worked, and the reasons why none of them were implemented in the real world. The second section, the introduction of neural networks, will follow the development from recurrent neural networks to transformer-based attention networks. We will explore the methodology and results of each one and conclude with the platform on which the chosen model is based.

\subsection{Pre-Neural Methods}


Most papers regarding general healthcare NLP can be divided into two topics: text classification and information extraction \cite{dong2022automated_11}. Classification can be split into three versions, each getting more complex: binary classification, where an instance is in one of two distinct categories (e.g., smoker or non-smoker); multi-class classification, where there are multiple categories, but an instance can still only be assigned to one class (e.g., current smoker, former smoker, non-smoker); and multi-label text classification \footnote{\url{https://huggingface.co/blog/Valerii-Knowledgator/multi-label-classification}}. This involves instances that can be associated with several different labels/categories simultaneously, such as discharge letters, in which each letter always contains multiple conditions. Automated medical coding is often identified as a multi-label text classification problem; however, some older attempts still utilise information extraction or a combination of methods from both topics. 
The first attempts at automated clinical coding were from around 1970, such as this 1973 study by \newcite{dinwoodie1973automatic_15} that utilises a ‘fruit machine’ methodology. This entails representing each significant word of a diagnosis with an associated code number and, like a fruit machine in a pub, the code is correct when a common code number appears for all words in the diagnosis. While this study returns impressive results with a correct coding rate of over 95\%, this is only done with a small collection of pre-coded morbidity data from 16 doctors around Scotland. Thus, the project will not scale up to the complex real-world scenario.

No real progress was then made for the next few decades. A 2010 literature review on clinical coding \cite{stanfill2010systematic} evaluated the results of 113 studies, the earliest being the above 1973 study, and concluded that while the systems hold promise, there has been no clear trend of improvement over time. Another interesting trend from this review is that, while no improvements had been made, researchers' interest was increasing, as all but 4 of the studies found were published after 1994. 
Examples of attempted innovation from this period include a study from \newcite{farkas2008automatic_17rule} focusing on rule-based automated radiology report coding. It uses a variation of multi-label classification that treats the assignment of each label as a separate task, as opposed to treating valid sets of labels as a single class. It then builds a rule-based expert system that operates on if-then codes through the ICD hierarchy. It uses decision trees (which recursively classify the data through conditions, similar to the rule-based system used to classify codes) to predict false positives, which occur when the model incorrectly predicts a positive outcome. It then uses a maximum entropy classifier to tackle false negatives, calculating each token's probability of a false negative. Both the decision tree and max entropy classifier worked to increase the micro-averaged $F_{\beta=1}$ scores by ~ 4\%, to 87.92\%.

While these rule-based solutions are very accurate for the specific types of documents they examine, they will not generalise well to new problems since they are domain-specific. For them to be feasible for real-world use, the rules would need to be extended to tens of thousands of codes and would require a substantial investment of time and expertise to be executed properly. Statistical approaches such as initial attempts from \newcite{mullenbach-etal-2018-explainable_CAML}, which utilised logistic regression (LR), and \newcite{perotte2014diagnosis_28}, which made use of Support Vector Machines (SVM), were attempted. However, the results on the full MIMIC database (shown in Figure \ref{fig:historical-methods-scores}) indected that they were also infeasible. Therefore, a different method had to be attempted: deep learning and neural networks. 

\subsection{Neural Networks and Attentions}

\begin{figure*}[!t]
\begin{center}
\centering
\includegraphics*[width=0.85\textwidth]{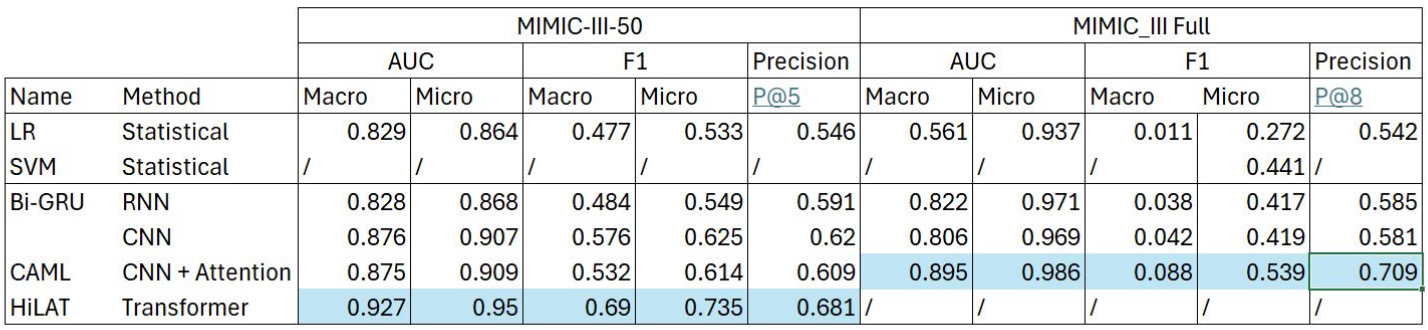}
\caption{Graph showing the AUC, F1 and Precision scores of the various methods explained in the Background section for both MIMIC-III-50 and MIMIC-III Full. Best scores for each category are highlighted in light blue. }
\label{fig:historical-methods-scores}
\end{center}
\end{figure*}

The general approach of deep learning in neural networks aims to map a complex function learned through the training data to match the information in the text to an appropriate set of medical codes \cite{dong2022automated_11}. 
Before any deep learning is completed, the common first step in these projects - aside from preprocessing - is to produce word embeddings for each token. Each embedding is a semantically meaningful mathematical representation, usually a vector, of the token designed so that tokens with similar meanings have similar vectors \cite{percha2021modern_18}. To compare the meaning of two words, one calculates the cosine similarity of their corresponding vectors. The most common method for doing this is ‘word2vec’, which operates on the assumption that words with similar meanings tend to occur in similar contexts. 
It uses either a continuous bag of words (CBOW) model that predicts the target words based on the context words (words surrounding the target word) or a skip-gram that predicts the context words based on the target words \cite{mikolov2013distributed_19}, both of which are examples of single-layer neural networks. 
A more advanced version of word2vec that strays from the standard embedding practice of one vector per word/token/document, is the development of bidirectional encoder representations from transformers (BERT) \cite{devlin-etal-2019-bert_21}. These are massive pre-trained language models that are too resource-intensive to be trained from scratch in most circumstances, however, models trained on a general corpus can be fine-tuned to meet specific needs (such as clinical text mining through transfer learning \cite{peng2019transfer_20}). Unfortunately, due to their size and complexity, they are not currently feasible to be trained on larger datasets without significant modification. 
The first successful deep learning attempts utilised recurrent neural networks (RNNs), with a focus on two specific types: Gated Recurrent Units (GRUs) and Long Short-Term Memory Networks (LSTMs). The project \cite{nigam2016applying_22} constructs an RNN with a single layer consisting of 20 time steps; with each time step, a normalised vector representing a patient note is submitted in a time sequential order (oldest to most recent). The activation (threshold) function is tanh, a mathematical operation applied to the weighted sum of inputs and biases in each neuron that introduces non-linearity into the network. There is a dropout rate of 0.1 that is applied to prevent overfitting during training, and a learning rate of 0.001 is used to determine how much the weights of the network are updated during each training iteration. Finally, the model uses cross-entropy loss as its sigmoid function, normalising the neuron's output to a value between 0 and 1. 

GRUs are implemented as recurrent units, where each unit contains a reset gate and an update gate, which allow the GRU to regulate the flow of information and selectively update its hidden state. They are computationally more efficient; however, they may be outperformed by LSTMs in tasks requiring long-range dependencies. LSTMs are built like GRUs but using three gates instead of two: an input gate, a forget gate, and an output gate. They are more powerful due to their additional gates and memory cells that allow them to better preserve information over time. 
Convolutional neural networks (CNNs) consist of convolution layers and pooling layers and are mainly used for image and video processing, however, if the text is manipulated and processed correctly, they can be very effective for text processing. For example, one of the most successful studies into automated medical coding is the 2018 project Convolutional Attention for Multi-Label Classification (CAML) \cite{mullenbach-etal-2018-explainable_CAML}, which utilised a CNN but swapped the pooling layer for an attention mechanism. This attention mechanism is applied to the data to identify relevant portions of the document for each code prediction, allowing it to selectively focus on and assign higher importance to the relevant words and phrases \cite{google2017attention}. Using attention mechanisms in this way also allows for enhanced interpretability. 
It provides insights into which parts of the document it made its predictions from, instead of just being put through a function as with previous methods. 

With attention comes transformer-based networks, and while attention networks are not exclusively transformer-based, 
transformers are exclusively attention-based \cite{google2017attention}. They rely solely on self-attention mechanisms, parallel processing the entire input sequence. This makes them more efficient for handling long sequences and allows for faster training and inference than more sequential models like RNNs. Transformers also allow for multi-head attention, an extension of the self-attention mechanism that allows the model to further parallelize the processing, enabling transformers to capture different aspects of the input data in parallel, allowing for more complex modelling of the relationships and patterns. This has recently been introduced into automated medical coding and, as demonstrated with HiLAT \cite{LIU2022104161_HiLAT}, it is already promising. However, due to the computational complexity of such a model, it has only been tested on the limited MIMIC-III-50 dataset. 


The table shown in Figure \ref{fig:historical-methods-scores} demonstrates automated coding techniques' slow but consistent progress. The highlighted segments represent the top performers in their respective categories. The transformer-based HiLAT model outperforms every other model in every metric when tested on the MIMIC-III-50 database. 
On the other hand, the CNN + attention-based model of CAML does the same when tested against all the models on the MIMIC-III Full database, while it is also the only model that can provide a level of explainability to its answers. 
These results indicate that an attention-based model is the preferred choice due to the superior results and their ability to provide explainability for their answers.

\subsection{The MIMIC-III Dataset}
\label{subsec_mimic}
In Clinical NLP, the first resource is the MIMIC-III dataset, which is the only publicly available mainstream English dataset with enough data to perform proper training. Additionally, most models that attempt to solve the automatic coding problem use this dataset. 

MIMIC-III \cite{johnson2016mimic} is a large, freely available database comprising de-identified health-related data associated with over forty thousand patients who stayed in critical care units of the Beth Israel Deaconess Medical Centre between 2001 – 2012 \footnote{\url{https://mimic.mit.edu/docs/iii/}}. The database is freely available to researchers worldwide, provided they have become a credentialed user of PhysioNet \cite{johnson2016mimic} and completed the required ‘Data or Specimens Only Research’ CITI training \footnote{\url{https://physionet.org/content/mimiciii/view-required-training/1.4/}} (Or another recognized course in protecting human research participants that includes HIPAA requirements).
All data in the MIMIC database has been deidentified per HIPAA (Health Insurance Portability and Accountability Act) standards. This ensures that all 18 listed identifying data elements, such as names, telephone numbers, and addresses, are removed. 
The only thing not removed are dates, which are shifted in a random but consistent manner to preserve intervals. Therefore, all dates occur between 2100-2200, but the time of day, day of the week, and approximate seasonality have been conserved. 

MIMIC is a relational database consisting of 26 tables containing different forms of data, from the patient’s clinical notes in NOTEEVENTS to extremely granular data such as the hourly documentation of patients’ heart rates. This makes it a vast and complex database to work with - however since we are only using the database for its clinical notes, only five tables are required:
\begin{itemize}
    \item NOTEEVENTS – Deidentified notes, including nursing and physician notes, ECG reports, imaging reports, and discharge summaries.
    \item DIAGNOSES\_ICD - Hospital-assigned diagnoses, coded using the International Statistical Classification of Diseases and Related Health Problems (ICD) system.
    \item PROCEDURES\_ICD - Patient procedures, coded using the International Statistical Classification of Diseases and Related Health Problems (ICD) system.
    \item D\_ICD\_DIAGNOSES - Dictionary of International Statistical Classification of Diseases and Related Health Problems (ICD) codes relating to diagnoses.
    \item D\_ICD\_PROCEDURES - Dictionary of International Statistical Classification of Diseases and Related Health Problems (ICD) codes relating to procedures.
\end{itemize}

This still leaves a lot of unnecessary data. For example, the NOTEEVENTS table contains CHARTTIME, CHARTDATE, and STORETIME, which are the time and date a note was charted and the time it was stored in the system. The notes in NOTEEVENTS vary in usefulness and format, with the type of note indicated in the DESCRIPTION column. Since all the medical coding projects that use MIMIC unanimously choose to use the discharge summaries as they contain the most potential codes per letter (15.9 labels per document).
We removed all the other types of notes. This was done by creating a new table that copied each line as long as the DESCRIPTION = ‘Discharge Summary’. The next step is to combine the data in separate tables into one table for easier access. 

\begin{figure*}[!t]
\begin{center}
\centering
\includegraphics*[width=0.85\textwidth]{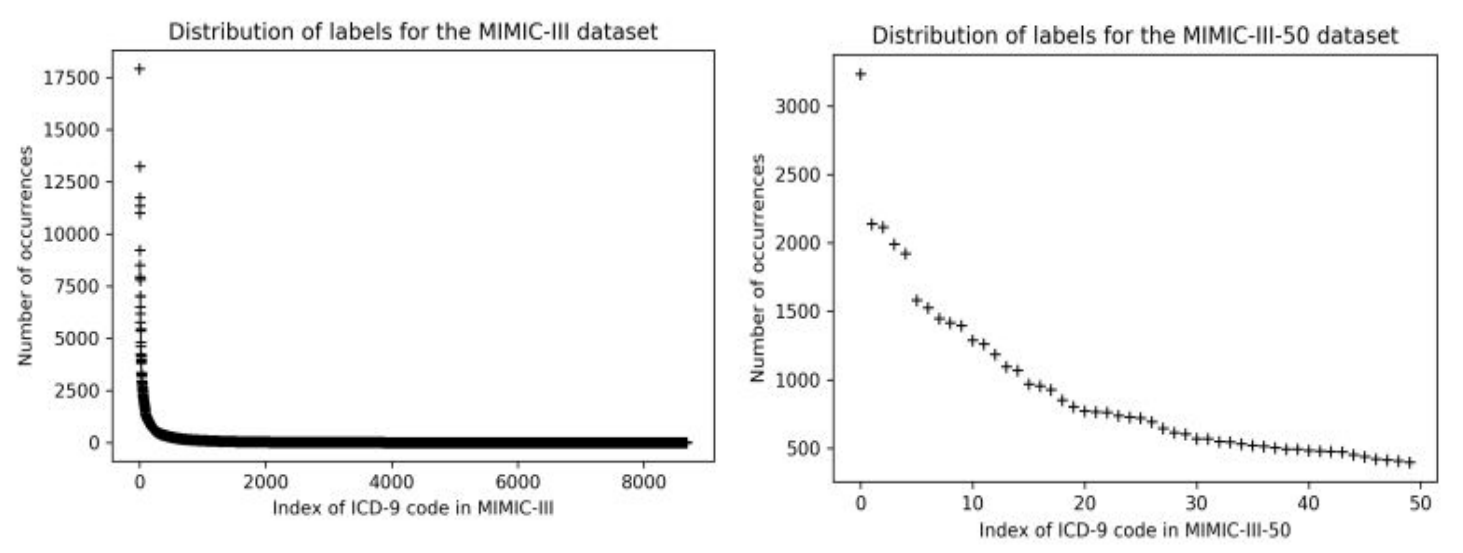}
\caption{Distribution of labels for the MIMIC-III and MIMIC-III-50 dataset
}
\label{fig:mimic-distribution_cut}
\end{center}
\end{figure*}

Another note on MIMIC is about its most popular subset, MIMIC-III-50, that contains only the notes and codes of the top 50 most frequently occurring codes (Table \ref{tab:mimic-iii-vs-50}). First occurring in CAML \cite{mullenbach-etal-2018-explainable_CAML}, MIMIIC-III-50 is often used as a proof-of-concept database for automatic medical coding projects due to it being significantly smaller (8,067 documents compared to 47,724) and with fewer labels (5.7 compared to 15.9 for MIMIC full), which means it takes less time and computational resources to train against. Projects like HiLAT \cite{LIU2022104161_HiLAT} that face challenges in accessing the necessary computing power for training their models have utilised the MIMIC-III-50 dataset to train on and achieve state-of-the-art results. 
The only issue with using MIMIC-III-50 is that, as Figure \ref{fig:mimic-distribution_cut} demonstrates, it doesn’t give the same opportunity to test models against a long tail distribution. 


A database that follows a long tail distribution is one where there are many data points that are not well-represented, and the majority of occurrences are concentrated around a few values at the ``head'' of the distribution \cite{10_zhang2023deepLong}. 
This accurately describes the MIMIC-III-Full database, where the top 105 codes make up 50\% of the total labels in the set, and there are 3,110 labels that have fewer than 5 examples \cite{nigam2016applying_22}, with 203 codes not appearing in any discharge summaries at all. Solving the long tail distribution of MIMIC is one of the key challenges that will need to be addressed by the potential models to be deployed.

\begin{table}[htb!]
    \centering
    \begin{tabular}{lll}
         & m.full  & mimic-iii 50 \\\hline
        training documents & 47,724 & 8,067\\
        Vocabulary size & 51,917 & 51,917\\
        Mean tokens per doc & 1,485 & 1,530\\
        Mean labels per doc & 15.9 & 5.7\\
        Total labels & 8,922 & 50\\\hline
    \end{tabular}
    \caption{Details regarding the discharge summaries in the MIMIC-III Full (m.full) and MIMIC-III-50 databases \cite{dong2021explainable_29HLAN}.
    }
    \label{tab:mimic-iii-vs-50}
\end{table}

\begin{figure*}[!t]
\begin{center}
\centering
\includegraphics*[width=0.85\textwidth]{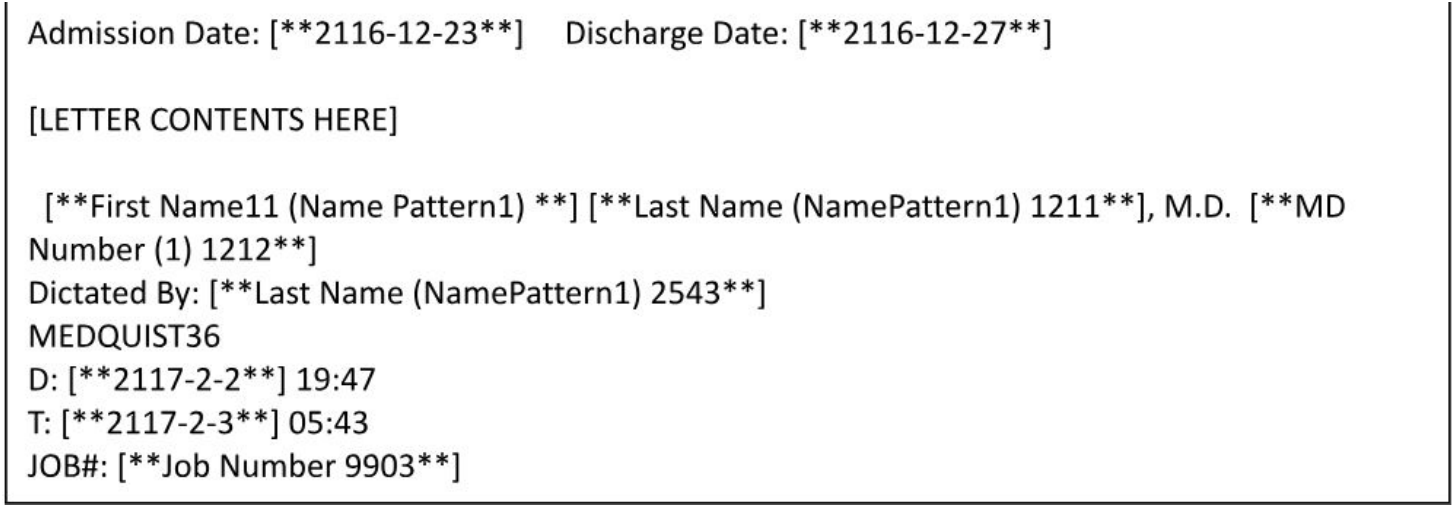}
\caption{Medical Letter Example }
\label{fig:letter-example-cropped}
\end{center}
\end{figure*}

\section{Model Selections}

We have selected three potential models and in this section each model will be evaluated, reviewing their results, methodology, and suitability for the study's needs, concluding with the chosen model. 

\subsection{Problem Formalisation}

Before each selected model is evaluated, the problem needs to be formally defined. Taking $\mathcal{X}$ as the collection of clinical notes and $\mathcal{Y}$ as the full set of labels (ICD-9 codes). Each instance $x_d \in  X$ is a word sequence of a document, $d$, and is associated with label set $y_d \subseteq Y$, where each $y_d$ can be represented as a $|Y|$ multi hot vector (a vector where multiple elements can have a value of 1, indicating multiple features/categories are present at the same time), $\overrightarrow{Yd} = [y_{d1}, y_{d2}, ... , y_{d|Y|}]$, and $y_{dl} \in (0,1)$ where $l$ indicates the $l’th$ label has been used for the $dth$ instance and 0 indicates irrelevance \cite{perotte2014diagnosis_28}. From this, the task of the models is to learn a complex function $f: \mathcal{X} \rightarrow \mathcal{Y}$ from the training set. 

All the chosen models use the same loss function, binary cross-entropy, and optimise it with L2 normalisation using the Adam (Adaptive Movement Estimation) optimiser \cite{kingma2014adam_31}. Loss functions are used in neural networks as a measure of how well the networks predictions match the true values of the training data, with binary cross entropy loss measuring the dissimilarity between the true binary labels and the predicted probability of the model. In the context of these models, L2 normalisation is used to avoid overfitting, which occurs when the model is trained so well on a particular dataset that it fails to generalise well to new, unseen data. To prevent this, penalty terms proportional to the magnitude of the vectors (Euclidean norm) are added, which penalise overly specific mappings and encourage the model to learn simpler, more generalised weight configurations. The Adam optimiser 
is a popular optimisation algorithm used to update the parameters of a neural network to minimise the loss function during training.

\subsection{Model-1: Convolutional Attention for Multi-Label Classification (CAML)}

\begin{figure}[!t]
\begin{center}
\centering
\includegraphics*[width=0.5\textwidth]{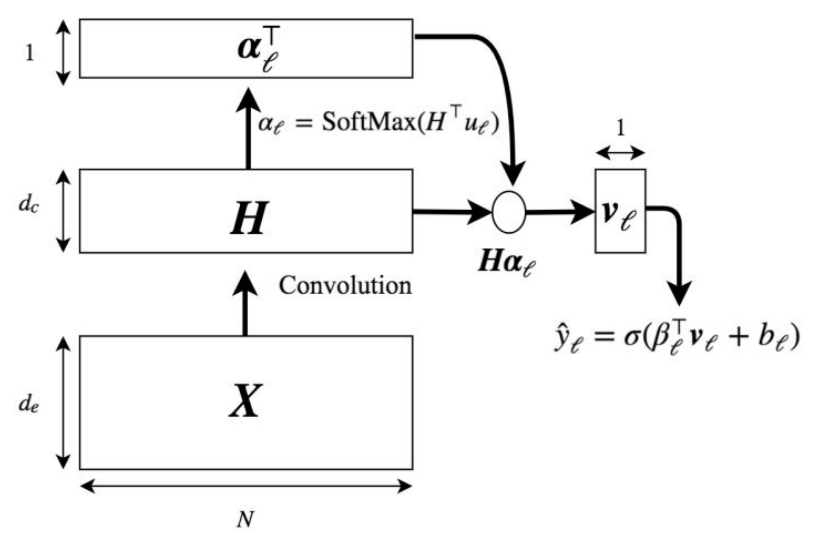}
\caption{The CAML architecture with per label attention shown for one label from
\cite{mullenbach-etal-2018-explainable_CAML}.
}
\label{fig:CAML-archi-cropped}
\end{center}
\end{figure}

CAML \cite{mullenbach-etal-2018-explainable_CAML} (Figure \ref{fig:CAML-archi-cropped}), as already mentioned in the background section, utilises a CNN based architecture but swaps the traditional pooling layer for an attention mechanism. The model starts by horizontally concatenating pretrained word embeddings into a matrix, X. A sliding window approach as is standard in CNNs is then applied to this matrix that computes an equation on each section of the matrix, resulting in the matrix H.

Next, the model applies a per-label attention mechanism. For each label, $l$, the matrix vector product is computed, and the result of this is passed through a SoftMax operator that essentially reduces the input values to the range [0,1] while ensuring that they sum up to 1 so they can be used as probabilities. This SoftMax operator returns the distribution over locations in the document in the form of attention vector $\alpha$. This attention vector is then used to compute vector representations for each label, $vl$. Finally, a probability is computed for label $l$ using another linear layer and sigmoid transformation to obtain the final label predictions $yl$. This normalisation process ensures that the probability of the label is normalised independently rather than normalising the probability distribution over all labels like the SoftMax operator does.

\begin{figure*}[!t]
\begin{center}
\centering
\includegraphics*[width=0.65\textwidth]{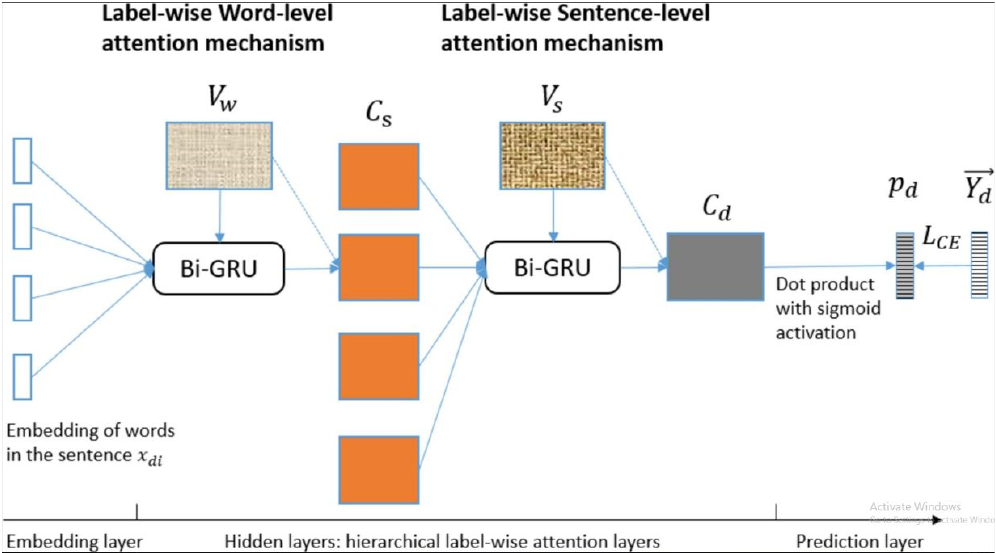}
\caption{The HLAN Model \cite{dong2021explainable_29HLAN}}
\label{fig:BiGRU-cropped}
\end{center}
\end{figure*}

\subsection{Model-2: Hierarchical Label Attention Network (HLAN)}

The HLAN model \cite{dong2021explainable_29HLAN} is built around providing explainability for its results, and consists of an embedding layer, the HLAN layers, and a prediction layer. The embedding layer converts each token in the sentence into a continuous vector where the word embedding algorithm word2vec 
returns the vector of word embeddings $x_{di}$.

The HLAN makes extended use of Gated Recurrent Units (GRU) to capture long-term dependencies. The GRU unit processes tokens one by one, generating a new hidden state for each token. At each hidden state, the GRU considers the previous tokens using a reset gate and an update gate. The GRU method implemented is known as Bi-GRU because it reads the sequence both forwards and backwards, concatenating the states at each step, to create a more complete representation.

The label wise word-level attention mechanism, which contains a context matrix ($V_w$) where each row $V_{wl}$, is the context vector to the corresponding label $y_l$. The attention score is calculated as a SoftMax function of the dot product similarity between the vector representation of the hidden layers from the Bi-GRU and the context vector for the same label. The sentence representation matrix $C_s$ is computed as the weighted average of all hidden state vectors $h^i$ for the label $y^i$.

The label-wise sentence-level attention mechanism is computed in much the same way, outputting sentence-level attention scores and the document representation matrix $C_d$. The prediction layer then utilises a label-wise, dot product projection with logistic sigmoid activation to model the probabilities of each label to each document. Finally, the binary cross entropy loss function is optimised with L2 normalisation and the Adam optimiser. 

The HLAN has an extra label embedding initialisation (denoted as +LE) that can be implemented in place of the normal embedding layer and functions by leveraging the complex semantic relations (how different elements are related to each other in terms of their meanings) among the ICD codes. The embedding works off for two correlated labels; one would expect the prediction of one label to impact the other for some notes, which is represented as giving each label representation corresponding weights. 
The HLAN model was based on the HAN model \cite{yang-etal-2016-hierarchical}, where the only difference between the two is that at the sentence and document level, HLAN utilises contextual matrices, whereas HAN uses contextual vectors. This means that while HLAN is more individually label-oriented, HAN still produces an attention visualisation for the whole document and the results are only slightly worse but reducing the computational complexity of training the model. 
HAN \cite{yang-etal-2016-hierarchical} model was originally proposed as ``Hierarchical Attention Networks for Document Classification''.

\subsection{Model-3: Multi-Hop Label-wise Attention (MHLAT)}

\begin{figure*}[!t]
\begin{center}
\centering
\includegraphics*[width=0.65\textwidth]{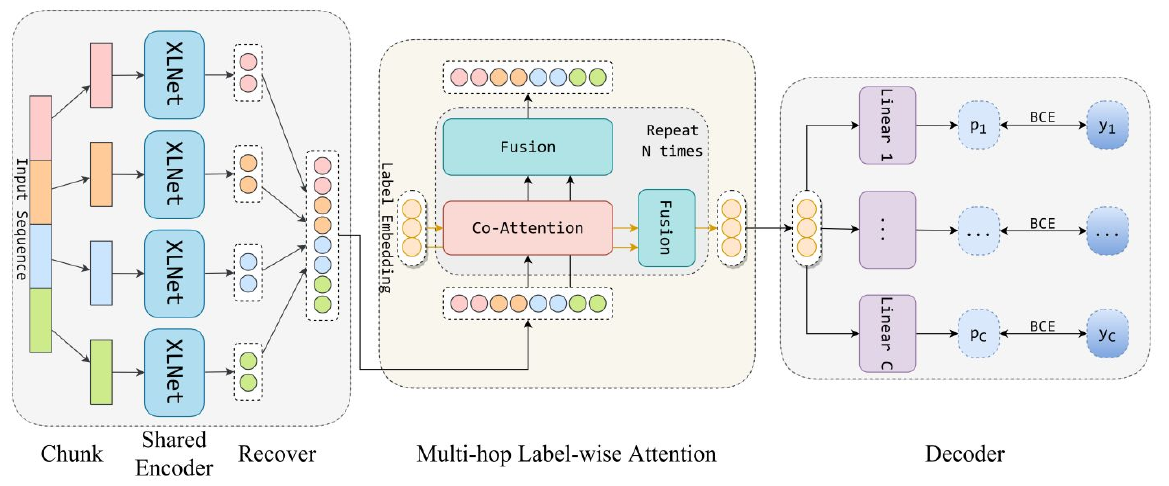}
\caption{The MHLAT model architecture \cite{duan2023mhlat_32icd-coding}.}
\label{fig:MHLAT-Model-cropped}
\end{center}
\end{figure*}

Much like HLAN, MHLAT \cite{duan2023mhlat_32icd-coding} is comprised of three main components: an input/encoder layer, MHLAT layer, and a decoder layer (Figure \ref{fig:MHLAT-Model-cropped}). 
It also utilises the same label-wise attention mechanism, however, that is where the similarities end. 
In the encoding layer, MHLAT first splits the text into chunks with 512 tokens per chunk. It then adopts the general domain pre-trained XLNet \cite{yang2019xlnet_33NLU} (similar to BERT but less computationally expensive), which is further trained on MIMIC, and then applied to every chunk. Each chunk from the text is then concatenated to form a global vector of the input text, H. 

While using label-wise attention through multiple passes is utilised for both HLAN and MHLAT, where HLAN uses multiple Bi-GRUs increasing the scope each time, MHLAT presents a ‘multi-hop’ approach. Initially, the label-wise attention is derived from matrices of the tokens of the input sentence from the encoder, followed by a ‘fusion’ operation that combines label-specific representations and label embeddings. A hop function is then defined that iteratively updates context information and label embeddings, which is then repeated. The decoding layer implements an independent linear layer for computing the label score and utilises the same binary cross entropy loss function as the other models.

\begin{figure*}[!t]
\begin{center}
\centering
\includegraphics*[width=0.65\textwidth]{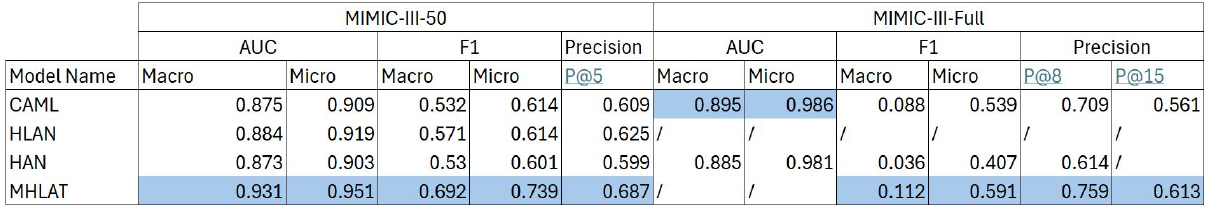}
\caption{The results of the above models on the MIMIC-III-50 and MIMIC-III-Full databases.}
\label{fig:4models-summary-cropped}
\end{center}
\end{figure*}

\subsection{Model summaries}

If going purely off results (given in Figure \ref{fig:4models-summary-cropped}), the MHLAT model returns state-of-the-art performance compared to the others in every metric it had resulted in. However, it is worth noting that the model, despite being attention-based, did not factor any type of \textit{explainability} into itself. 
As mentioned in the motivations, we want to explore some level of interpretability of coding models, otherwise, the professionals (clinicians) using them would have no way to verify the results and build trust. 

Looking at the results of the remaining models, it is clear that HLAN performs better than CAML, which in turn performs better than HAN. 
However, the objective of the project was to prioritise explainability in the results, which made HLAN/HAN the ideal model as despite a slight reduction in performance for the MIMIC Full dataset.
The enhanced \textbf{interpretability} in its answers justifies its use, especially in domains such as medical coding where transparency and understanding of the models’ decisions are crucial. 

\begin{figure*}[!t]
\begin{center}
\centering
\includegraphics*[width=0.85\textwidth]{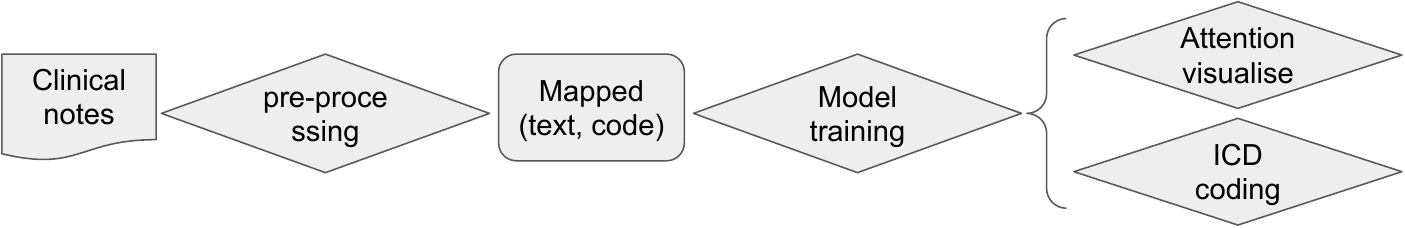}
\caption{Data Processing for Model Learning Pipeline. 
}
\label{fig:Jamie-pipelines-dataPro-ModelTrain}
\end{center}
\end{figure*}

\begin{figure*}[!t]
\begin{center}
\centering
\includegraphics*[width=0.85\textwidth]{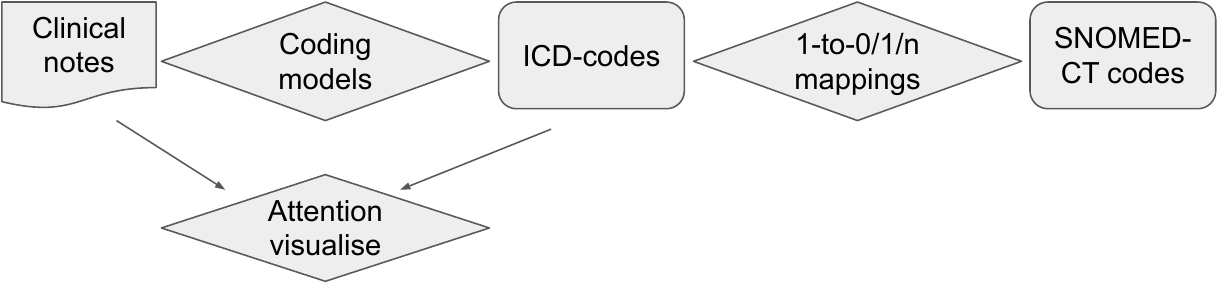}
\caption{Model Deployment Pipleline with ICD Coding Visuralisation and Mapping to SNOMED CT.
}
\label{fig:Jamie-pipelines-deploy-cropped}
\end{center}
\end{figure*}

\section{Coding with Explainability}

The goal of this study is to develop a program that could attempt to fulfill the investigation aims, that being to produce {SNOMED} codes and visualisation, and could then be utilised to evaluate a comparable system being implemented in the real setting, such as NHS UK. The program was implemented in \textit{Python 3.8} using the \textbf{TensorFlow} framework and leverages the \textbf{HAN} model 
to \textbf{predict ICD codes}, converts these codes to \textbf{SNOMED}, and provides \textbf{visualised attention scores} for each document.

\subsection{data processing and ICD coding}

The preprocessing (Figure \ref{fig:Jamie-pipelines-dataPro-ModelTrain}) takes three of the tables from MIMIC described in Section \ref{subsec_mimic}, NOTEEVENTS, PROCEDURES\_ICD, and DIAGNOSES\_ICD, and combines them into one table, notes\_labeled, with the schema SUBJECT\_ID, HADM\_ID, TEXT, LABELS where:

\begin{itemize}
    \item SUBJECT\_ID – identifier unique to a patient, found in NOTEEVENTS. 
    \item HADM\_ID – identifier unique to a hospital stay, found in NOTEEVENTS.
    \item TEXT – The free text of the document. There can be multiple documents with the same HADM\_ID. Found in NOTEEVENTS.
    \item LABLES – ICD\_9 labels professionally assigned and stored in sequence order in either DIAGNOSES\_ICD or PROCEDURES\_ICD, depending on if they were diagnoses or procedures.

\end{itemize}

This is accomplished by first concatenating both _ICD tables into one table of codes, ALL_CODES. In the next step it preprocesses the raw TEXT from NOTEVENTS, removing tokens that contain no alphabetic characters (i.e., removing 500 but not 500mg), removing white space, and lowercasing all tokens. The processed text is stored in the disch_full table, which is then joined on the HADM_ID of each line to the ALL_CODES table to form the notes_labeled table. 

The code then generates the MIMIC_III_50 database by iterating through the notes_labeled file, counting the occurrences of each code, and saving the HADM\_IDs to 50\_hadm_ids and the codes to TOP\_50_CODES. Both the standard notes_labeled and the dev\_50 tables are split 90/10 to train/test respectively and stored in the train/test version of their tables. 

When attempting to train the HLAN model on the full MIMIC dataset, the system that it was being trained on (our local PC) did not have sufficient memory, therefore the HAN model \cite{yang-etal-2016-hierarchical} was used instead. This model did not need to be trained as the pretrained model could be downloaded from the GitHub. 

There is now a working model that took a text document as input and outputted an attention visualisation in Excel and a list of predicted codes in the console. 

\subsection{Entity Linking to SNOMED}

Now with a working model, the next step is to map the ICD codes to SNOMED (Figure \ref{fig:Jamie-pipelines-deploy-cropped}). The map \footnote{\url{https://www.nlm.nih.gov/research/umls/mapping_projects/icd9cm_to_snomedct.html}} was originally created for the Unified Medical Language System (UMLS) to facilitate the translation of legacy data still coded in ICD-9 to SNOMED CT codes. Therefore, it is perfect for the project's needs. It does contain multiple columns of data that are not required, mainly usage statistics, however, these can just be ignored. 
The 202212 most recent release of the map was implemented by UMLS and is split up into two tab-delimited value files with the same file structure; one for one-to-one mappings, and one for one-to-many mappings. The one-to-one mapping contains 7,596 mappings (64.1\% of ICD-9 codes), with each line in the file being a separate mapping. For example, the ICD code 427.31 (Atrial Fibrillation) maps directly to the SNOMED code 49436004 (Atrial Fibrillation (disorder)). The one-to-many file contains 3,495 mappings (29.5\% of ICD-9 codes), with the mapping being one ICD code to multiple SNOMED codes. The file is set out as one-to-one maps, with the one ICD code being repeated for each of the many SNOMED codes, for example:
\begin{itemize}
    \item 719.46 – Pain in joint, lower leg $|$ 202489000 – Tibiofibular joint pain
    \item 719.46 – Pain in joint, lower leg | 239733006 – Anterior knee pain
    \item 719.46 – Pain in joint, lower leg | 299372009 – Tenderness of knee joint
\end{itemize}

This was implemented by first loading the one-to-one map into a dictionary, then iterating through the predicted_codes list.
At each iteration (new ICD code) the program checks to see if the ICD code is in the one-to-one map. 
If it is, the associated SNOMED code and FSN (fully specified name) are outputted; if not, the one-to-many map is loaded as a dictionary. 

The program searches for the ICD code in the one-to-many dictionary, and if found, it outputs all the SNOMED codes related to the ICD code. This is done so that even if the program cannot find a direct mapping, it can at least provide the user with potential options. If an ICD code cannot be found in any mappings, the system will print the ICD code description from either D\_ICD\_DIAGNOSES or D\_ICD\_PROCEDURES. There are only a few cases, approximately 6.4\% of the ICD codes, where there are no mappings available. This usually occurs with catch-all NEC (not elsewhere classified) ICD codes, such as 480.8 - Pneumonia due to other virus not elsewhere classified, for which SNOMED has no alternative mappings available.

\begin{figure*}[!t]
\begin{center}
\centering
\includegraphics*[width=0.99\textwidth]{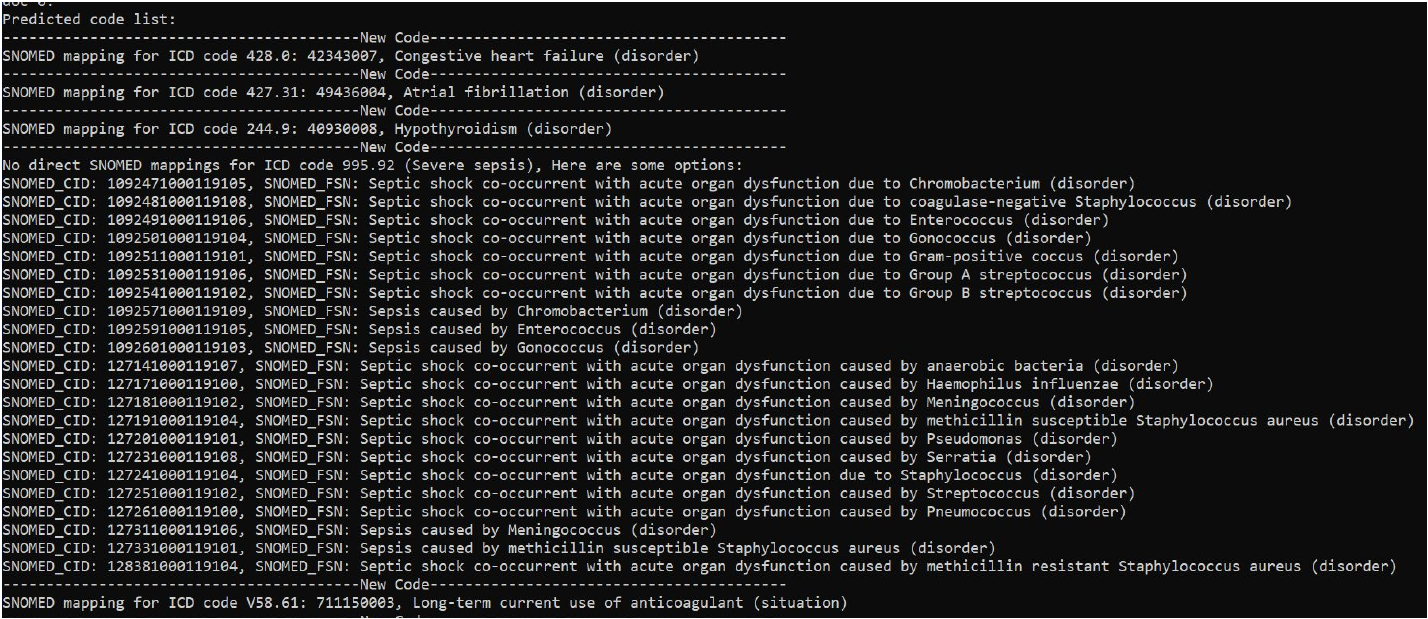}
\caption{Examples with the program returning 1-to-1 and 1-to-many ICD – SNOMED mappings.}
\label{fig:ICD-SNOMED-mapping-cropped}
\end{center}
\end{figure*}

\begin{figure*}[!t]
\begin{center}
\centering
\includegraphics*[width=0.99\textwidth]{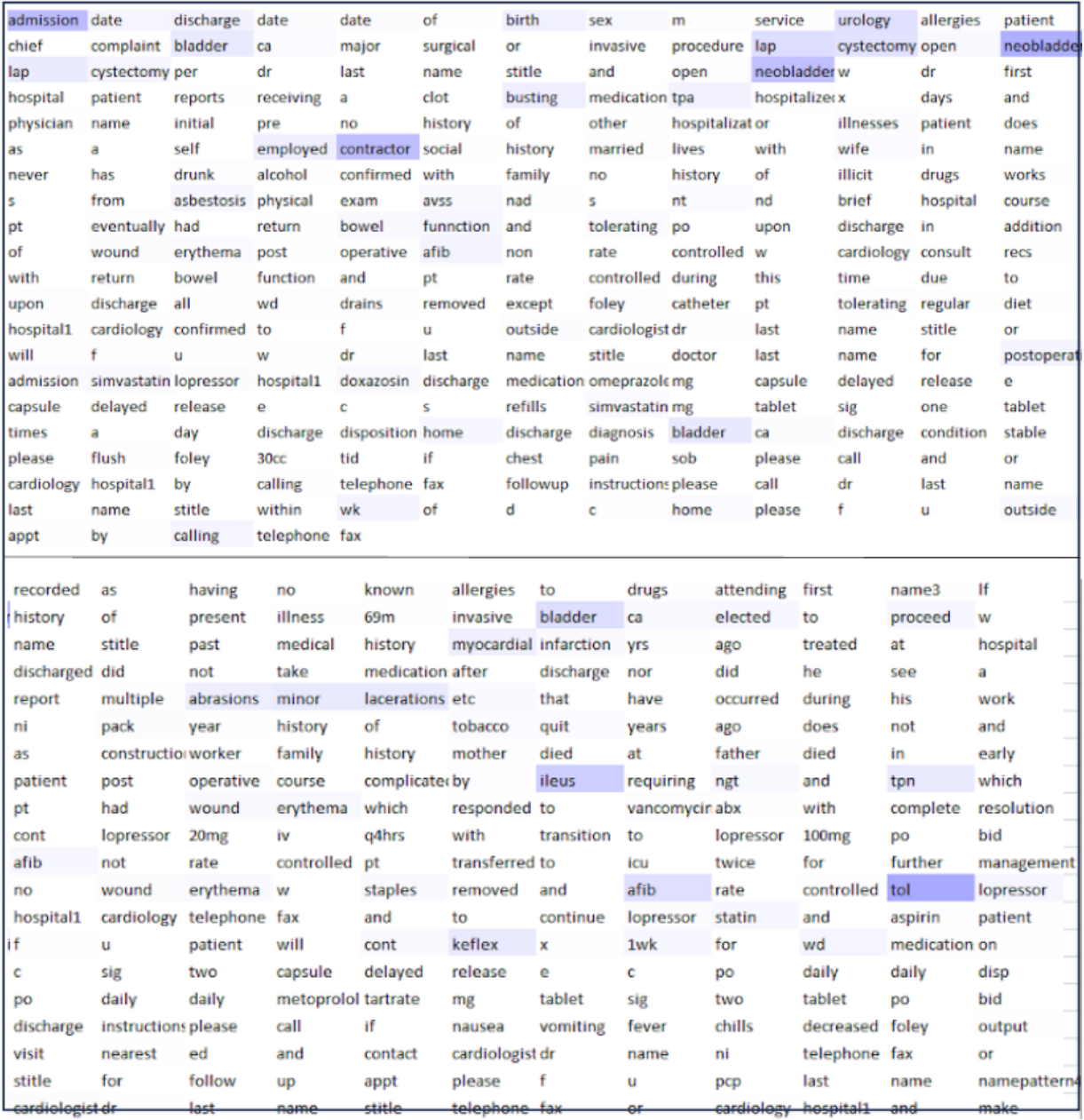}
\caption{Attention visualisation for the results of mapping.
The more blue something is highlighted, the more it was used to calculate the mapping. }
\label{fig:attention-visual-cropped}
\end{center}
\end{figure*}

After all these steps, the project now takes notes as input through a text document, processes them using the HAN model, and calculates the attention levels of the ICD codes. The program then converts the ICD codes into SNOMED codes with as many 1-to-1 mappings as it can find, outputting that to the console (Figure \ref{fig:ICD-SNOMED-mapping-cropped}). 
Finally, the attention visualisation is exported into Excel (Figure \ref{fig:attention-visual-cropped}) which shows each word in the file and highlights it in a shade of blue. The deeper the blue highlight, the greater the weight that word had when calculating the ICD codes. 
The visualisation displayed in Figure \ref{fig:attention-visual-cropped} is split up halfway down for ease of viewing. In reality, the left-hand side of the upper picture and the right-hand side of the lower picture are joined next to each other. 


\subsection{Evaluations Setups}

The experiments are evaluated in two ways – first, the model is tested against the standard testing scores of micro/macro F1 and precision. Second, the implementation of SNOMED mapping is also considered, calculating the percentage of codes it can predict/give options for. 

To accurately test the model, data had to be gathered by running the model against MIMIC discharge summaries from the test files. This was accomplished by randomly selecting 100 notes from the test_full file
(refer to sample size and model confidence by \newcite{gladkoff-etal-2022-measuring}). We then ran each set of notes through the model and put it through a program
that returned the true and false positives, as well as the false negatives from the results by comparing the labels generated by the model to the true labels in the file, where: 

\begin{itemize}
    \item True Positives – when the model predicts a label, and it is correct.
    \item False Positives – when the model predicts a label, but it is incorrect. 
    \item False Negatives – when the model doesn’t predict a label even though there is a correct label.

\end{itemize}

Now that these values were generated, the model was tested against the same metrics that have been used in all the models previously. 
\begin{itemize}
    \item Recall - measures how often a model correctly identifies positive instances (true positives) from all the actual positive samples in the dataset \footnote{\url{https://www.evidentlyai.com/classification-metrics/accuracy-precision-recall}} and is calculated by dividing the number of true positives by the number of positive instances (true positives + false negatives).
    \item Precision – measures how often a model correctly predicts the positive class, calculated by dividing the number of correct positive predictions (true positives) by the total number of instances the model picked as positive (both true and false positives). The precision results from earlier models were with P@5, P@8, or P@15, which means measuring the proportion of relevant items within the top 5, 8, or 15 items retrieved by the system. 
    \item F1 Score – Calculated as the harmonic mean of the precision and recall scores, therefore, encouraging similar values for both precision and recall. The more the precision and recall deviate from each other, the worse the score.
    \item Macro F1 score - is an average of the F1 scores obtained, representing the average performance of the model across all classes (each class having the same weight).
    \item Micro F1 score - computes a global average F1 score by counting the sums of the true positives, false negatives, and false positives and then putting those into the normal F1 equation. It essentially computes the proportion of correctly classified observations out of all observations (each token having the same weight).
\end{itemize}

Aside from gathering these results, the other data collected was that of the SNOMED scores. This was gathered when running the same tests to find the other values, and each returned SNOMED score could be grouped into one of 4 categories:
\begin{itemize}
    \item 1-to-1 – The ICD to SNOMED code was a one-to-one match
    \item 1-to-M – The ICD to SNOMED code was a one-to-many match
    \item No Map – No ICD to SNOMED map was found.
    \item No DESC – There was no description found associated with the ICD codes in the D\_DIAGNOSES\_ICD MIMIC file. This was a rare valid return due to the formatting of the D\_DIAGNOSES\_ICD file. 

\end{itemize}

\begin{figure*}[!t]
\begin{center}
\centering
\includegraphics*[width=0.99\textwidth]{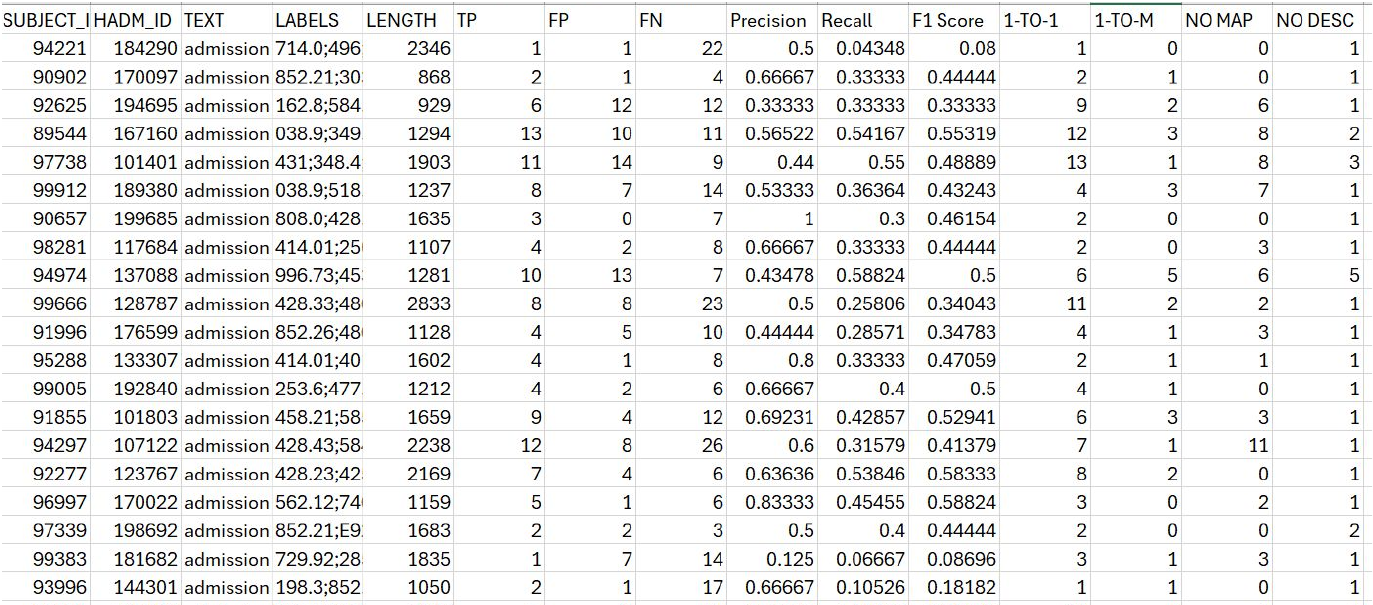}
\caption{The evaluation results of the first 20 documents tested (full results in appendix).}
\label{fig:eval-short-cropped}
\end{center}
\end{figure*}

\subsection{Evaluation Results}

\subsubsection{ICD Coding Evaluation}
For ICD coding evaluations, the first 20 documents tested were listed in Figure \ref{fig:eval-short-cropped}, with the full list in Appendix.


The combined results of all the tests (Table \ref{tab:HAN-vs-our-test}) were then calculated, returning the macro F1 as 0.041 (compared to 0.036 from previous HAN tests) and the micro F1 as 0.403 (compared to 0.407 from previous HAN tests). 
The similarity to the previous results demonstrates that the model was functioning as intended, so although the results weren’t state of the art, they were what was expected. 
The same can be said for precision, which We calculated using the first 15 values returned, otherwise known as P@15 (the same as previous tests), to get a precision of 0.599 (compared to 0.613). 

While these results aren’t the same as the previous HAN model testing, this is to be expected as only 100 documents were tested. This means that if there were outliers, they had a greater effect on the overall results, and the more documents that were tested, the closer to the actual values the results will become. 

\begin{table}[htb!]
    \centering
    \begin{tabular}{cccc}
        Models & Precision@15  & Macro F1 & Micro F1 \\\hline
        HAN-our & 0.599 & 0.041 & 0.403 \\
        HAN-ori & 0.613 & 0.036 & 0.407 \\\hline
    \end{tabular}
    \caption{Combined results comparing our HAN testing against the original HAN results.}
    \label{tab:HAN-vs-our-test}
\end{table}

\begin{table}[htb!]
    \centering
    \begin{tabular}{ccccc}
         & 1-to-1  & 1-to-M & No Map & No Desc \\\hline
        Total & 446 & 117 & 263 & 17\\
        \% Total & 52.91\% & 13.88\% & 31.20\% & 2.02\% \\ \hline
    \end{tabular}
    \caption{Results of the SNOMED mappings.}
    \label{tab:SNOMED0-mapping-eval}
\end{table}

\subsubsection{SNOMED Mapping Evaluation}
Regarding the SNOMED mapping, from the individual results (shown in Figure \ref{fig:eval-short-cropped}), each row was summed, with 100 subtracted from the No DESC value to ensure that the error of the program producing a No DESC result at the end of each document was not considered in the total. 
From this, a 1-to-1 map is displayed 52.91\% of the time, and a 1-to-many map is displayed 13.88\%, which means the program successfully mapped to SNOMED on 66.79\% of attempts. 

The unexpected result in this situation is the significant amount of ‘no maps’ returned. This is due to differing versions of ICD-9 codes utilised, as MIMIC uses the standard ICD-9 coding, but the mapping uses ICD-9-CM, the clinical modification used for morbidity coding. 
This means that there will be codes in one version that are not featured in the other, and unfortunately, there is not much that can be done to resolve this aside from creating a new mapping. 

Even when returning a ‘no map’, the program still returns the description of the ICD code which is useful information for the user. Therefore, this implementation returns a useful response for 97.98\% of attempted codes.

\section{Conclusions and Future Work}
This study aimed to compare existing coding methods and produce a model that automatically assigns labels to medical texts and gives an explainable outcome, to explore how this investigation can be implemented in real practice, e.g. NHS UK.
High ethical standards were maintained during the project considering the field of study. 
As outcomes, the model does automatically assign labels to the medical texts utilising a pre-trained HAN model that emphasises \textit{interpretability} in its outcomes, producing a document explaining how it reached its decisions. 
The project also explores the potential of integrating a similar system into a real setting, 
utilising mappings to SNOMED as well as having a medical professional give feedback throughout the development of the system and evaluate the results of the final program (Appendix for human evaluations). 

Regarding future works specifically for real applications,
we believe that for a project like this to be viable, a new dataset needs to be created that more accurately represents the data the model is going to come across. 
Using discharge summaries from MIMIC to train the model and then expecting it to perform on completely different data is infeasible; no matter how complex the model is and how good it gets at zero-shot learning, etc., it will only ever be good at modelling data that is similar to the data it's trained against. 
Making a new database would also eliminate the need to map between coding standards, as making a new database specifically for use cases, e.g. NHS UK, means it can be mapped to SNOMED by default. 
Another direction is that we can deploy some SOTA medication and treatment extraction tools for richer annotation of clinical data, such as recent work by \newcite{Belkadi2023etal_PLM4clinicalNER,Tu2023etal_MedTem}.


From a more general perspective, automated medical coding as a problem seems to be advancing towards transformer-based solutions in both the full modelling like MHLAT and word embeddings with BERT. This technology shows definite promise with its results against MIMIC-III-50, with its only limit being the computational feasibility of training such a complex model.

\section*{Limitations}
After our first meeting, the external stakeholder created a simplified mock-up of the NHS Electronic Health Record (EHR) system to store patient information \footnote{\url{https://github.com/furbrain/SimpleEHR}}. The system integrated the SNOMED codes into the EHR utilizing the SNOMED terminology service Hermes \footnote{\url{https://github.com/wardle/hermes} Hermes : terminology tools, library and microservice.}. 
Since one of the objectives of the project was to demonstrate how it could be implemented into the wider NHS system, and creating a mock-up of the EHR was deemed as a good starting point.

Unfortunately, there were issues getting Hermes (more specifically the Hermes docker file) to function on a Windows PC, but these issues did not persist on the university virtual machines (VM), therefore the project was moved on to the Linux-based VMs. 
Doing this had its own problems, as we no longer had permissions to ‘sudo install’ any of the Python libraries required to run Hermes. 
To solve this, a custom text-based VM had to be created with all the permissions needed to run Hermes. There were \textit{access} problems regarding this VM with incorrect SSH keys, but once this was fixed a Hermes terminology server was successfully set up on the VM. 

Gaining access to MIMIC-III required the completion of two CITI training modules; Data and Specimens only research, and Conflicts of Interest (Both in Appendix). After this, our PyhsioNet account (PhysioNet is a repository of medical data, and where MIMIC is available to download) became credentialed and, therefore, gained access to the full MIMIC dataset.  

Unfortunately, the custom VM did not have enough \textit{space} for the full MIMIC dataset. Therefore, the dataset had to be downloaded onto our personal Windows PC without the working Hermes server and restart the project from there. From here preprocessing could begin to make MIMIC and the HLAN compatible.

\section*{Acknowledgements}
We thank the external stakeholder (a Local GP) for the support, feedback, and human evaluation during this project.
LH and GN are grateful for the grant “Integrating
hospital outpatient letters into the healthcare data space” (EP/V047949/1;
funder: UKRI/EPSRC).

\bibliography{anthology,custom}
\bibliographystyle{acl_natbib}
\newpage
\appendix

\section*{Appendix}
\label{app-all}

\section{Study Context}
This paper explores the potential of replacing the time-consuming process of manually coding letters with a program that automatically assigns codes to letters. For the program to be of any value to its intended users, the external stakeholder (who is a local GP and has an interest in programming) stated that the output should be explainable. 
This would allow the users to verify the results if unsure and increase the trust between them and the system.  The stakeholder also stated that ideally the system would be easily implemented into the wider NHS systems, so the system can store and link the codes and letters to the patients they are about. This would allow the program to utilise previous letters about the patient to aid with the coding.  

Due to the program being oriented around the inherently personal topic of healthcare, ethics approval to gain access to the resources required would always be important. we had to gain access to MIMIC-III (Medical Information Mart for Intensive Care) which is a free database comprised of deidentified healthcare data, as well as the UK and US versions of SNOMED-CT and access to the UMLS ICD-9 to SNOMED-CT maps from the NIH. 
The MIMIC database had to be pre-processed to train the HLAN (Hierarchical Label Attention Network) system that generated the ICD-9 label predictions. These label predictions had to be mapped to SNOMED-CT terminology codes, and the label predictions exported in a user-friendly and readable manner. 

The external stakeholder will evaluate this, and tests will be created to validate the results already generated by the HLAN and see if mapping to SNOMED affects them.

The following training was conducted for the good practice:
\begin{itemize}
    \item CITI training \footnote{\url{https://physionet.org/about/citi-course/}}: collaborative institutional training initiative (CITI Program)
    \item Massachusetts institute of technology affiliates
    \item Curriculum group: Human Research 
    \item Course Learner Group: Data or Specimens Only Research
\end{itemize}

\section{Human Evaluation Insights}
The second method of our evaluations is to allow the stakeholder to try and code some example real-world scenario letters. To evaluate this program, we will collect the results of the program coding those letters, as well as the stakeholders verbal feedback on how this would fit within the NHS.

To complete the stakeholder evaluation, the external stakeholder prepared six example letters containing a mix of common and uncommon diseases/procedures that they would come across in their everyday work. The letters included sections designed to test the system, such as the example letter below signed by ‘Dr xxx xxx’:

Dear Dr xxx,
Thank you for sending xxx to me. I agree that I think she has quite bad psoriasis; I will refer her for phototherapy.
Yours Sincerely,
Dr xxx xxx

\begin{figure*}[!t]
\begin{center}
\centering
\includegraphics*[width=0.99\textwidth]{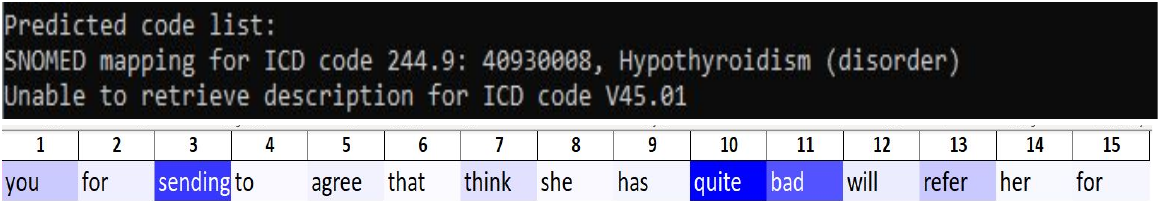}
\caption{Codes returned and the attention map presented to the external stakeholder for the example letter. It should be noted that the code V45.01 = cardiac pacemaker in situ.}
\label{fig:human-eval-piece-cropped}
\end{center}
\end{figure*}

The letters were processed with the model, and the predicted codes and their attention maps were shown to the stakeholder (the other letters are contained in Appendix \ref{appendix_letters}). Unfortunately, the results on almost all the letters were disappointing. With the letter above, the correct codes would be 9104002—psoriasis and either 31394004—light therapy, which is the parent to all forms of phototherapy, or 428545002—phototherapy of skin as the more specific result. The model returned the results and attention map shown in Figure \ref{fig:human-eval-piece-cropped}.

\begin{figure*}[!t]
\begin{center}
\centering
\includegraphics*[width=0.99\textwidth]{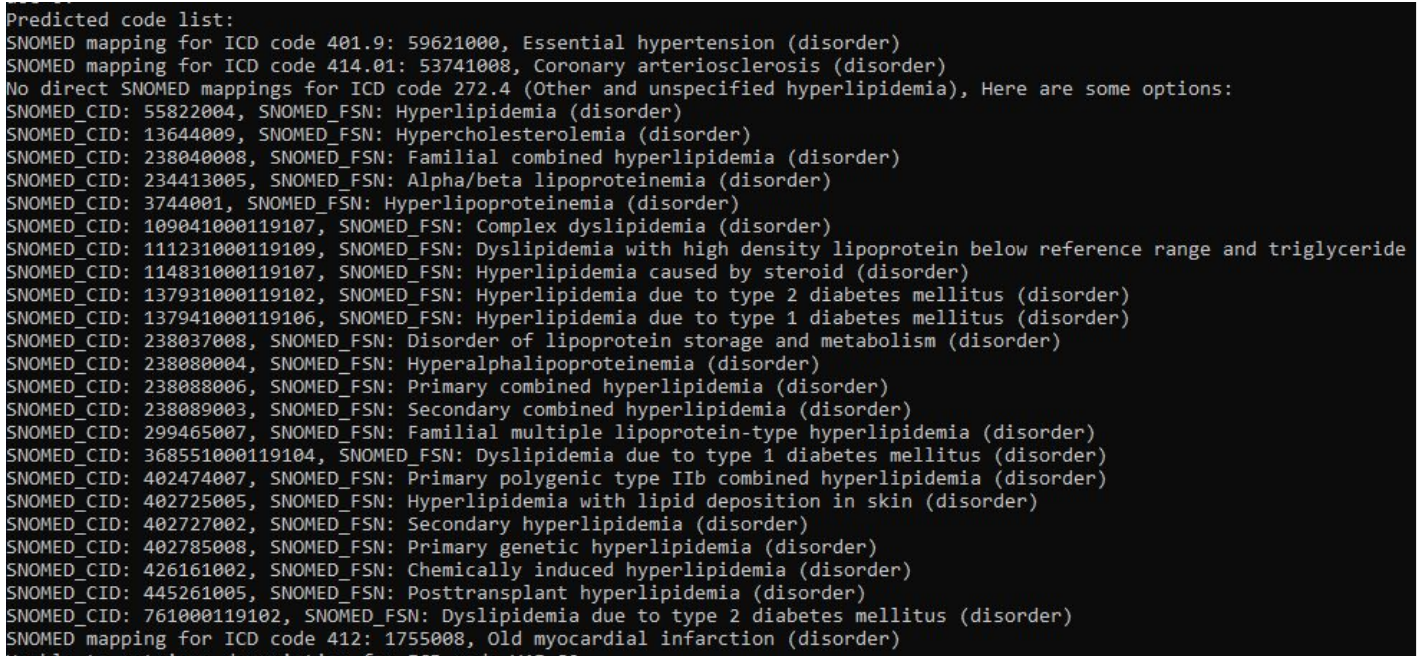}
\caption{
Result given to the external stakeholder with examples of direct and indirect SNOMED mappings.}
\label{fig:SNOMED-map-2stakeholder-cropped}
\end{center}
\end{figure*}

With these results, not only were the predicted codes incorrect but the attention maps were also both wrong and removing words. 
This \textit{did not} happen with any of the MIMIC discharge summaries, which, even when the codes were wrong, at least specified where in the letter the codes were found (as demonstrated in Figure \ref{fig:SNOMED-map-2stakeholder-cropped}). 

There was one letter where the result was correct; the letter stated, ‘I reviewed xxx following his PCA - this has indeed shown a MI which is clearly causing LVF, as evidenced by his raised BNP. We will proceed to a CABG’, where, in this case, LVF = left ventricular failure and CABG = coronary artery bypass graft. The model returned with 42343007 - congestive heart failure, which the external stakeholder identified as a perfect match for LVH, and the procedure ‘continuous invasive mechanical ventilation for less than 96 consecutive hours’, which, although oddly specific, does occur during a CABG. 

Since using the pre-prepared letters didn’t give the system a chance to demonstrate how it returns the codes, the external stakeholder was also given the codes returned from a MIMIC discharge summary (Figure \ref{fig:SNOMED-map-2stakeholder-cropped}) that showed codes with direct and indirect SNOMED mappings. Regarding this, they stated that with a good enough accuracy of coding, the solution would genuinely be useful for medical coding, with their only critique being that when there is no direct mapping, usually the least specific (parent in the hierarchy – in the example in Figure \ref{fig:SNOMED-map-2stakeholder-cropped} that would be 55822004 - Hyperlipidaemia) should be used. 

From these results conclusions can be made looking at the issues from two angles. The first is that, despite the best efforts of the model, it has succumbed to overfitting with the MIMIC discharge summaries, leading to it not properly functioning when given data that doesn’t resemble said discharge summaries. 

The other conclusion is that the MIMIC database simply isn’t representative enough of what this project aims to code. The model is only trained using discharge summaries, which are long and detailed documents, but more importantly, they only contain diseases/procedures that would require hospitalisation. This also explains why the model successfully predicted heart failure – a serious condition that presumably would have been included in multiple discharge summaries – but didn’t detect the other letters (included in Appendix \ref{appendix_letters}) about less serious diseases such as ear infection, headaches, and psoriasis. 

A note on this conclusion is that the final letter that describes ‘Waldenström’s Macroglobulinemia’ – a rare form of blood cancer - returned no mappings despite it being something with potential for hospitalisation. This was still the case when we changed it to its other well-known name, lymphoplasmacytic lymphoma. 

Finally, the stakeholder stated that another thing to be added to make it truly useful would be that it implements the whole of the SNOMED terminology, not just the \textit{diagnoses} and \textit{procedures}. Using MIMIC data, the models can only be trained on ICD-9 codes, which as described earlier only contain diagnoses and procedures. SNOMED also has hierarchies for medicines, tests, organisms, and substances that also need coding.

\begin{figure*}[!t]
\begin{center}
\centering
\includegraphics*[width=0.99\textwidth]{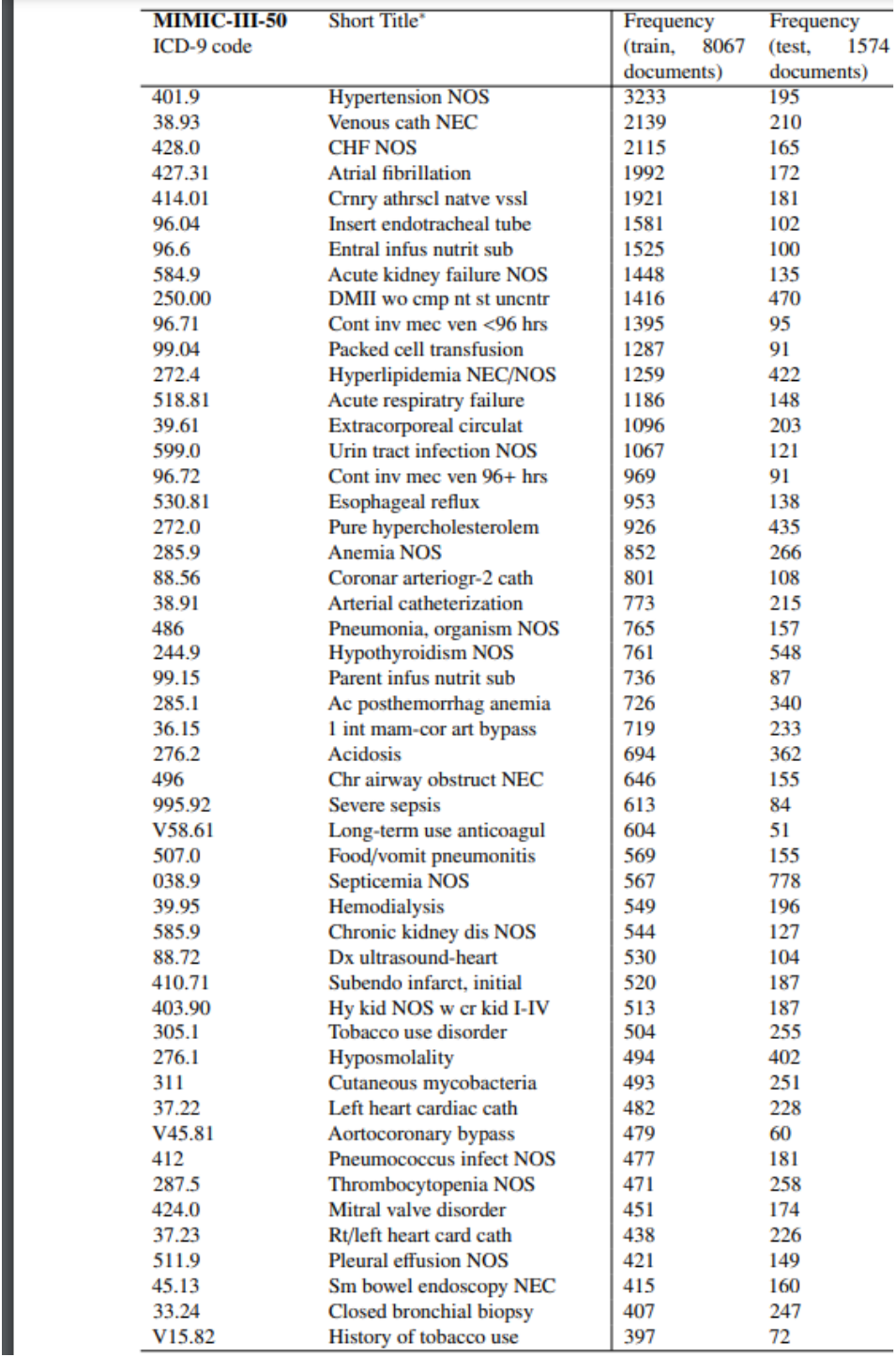}
\caption{
MIMIC-III-50 Table of Included Codes and corresponding short title of ICD-9 code \cite{dong2021explainable_29HLAN}
}
\label{fig:MIMIC-III-50Codes-cropped}
\end{center}
\end{figure*}




\section{Implementation Details}

Implementing the HAN model came with surprisingly few difficulties considering its complexity and the previous issues with everything in the project so far. It required Python 3.8 instead of 3.6 and TensorFlow 1 instead of PyTorch like CAML. A note on TensorFlow 1 - The only version available for download is TensorFlow 1.15, deprecated from TensorFlow 2.0.0 and installed through the TensorFlow Hub onto an Anaconda (conda) virtual environment.

To preprocess the data so that it is in the format expected for the \textbf{HLAN} model to train/test, it requires the same preprocessing as CAML. There were some issues running this as some of the Python libraries, more specifically the versions of NumPy, SciPy, and Scikit-Learn in the requirements list, kept throwing errors about each other’s versions on installation. 
This was fixed by doing a clean install of Python 3.6 in a virtual environment, and this virtual environment was where the CAML preprocessing script was run \footnote{\url{https://github.com/jamesmullenbach/caml-mimic}}. In this virtual environment there were problems running Jupyter Notebook, but to fix this, the code was copied from the notebook into a regular Python file that did what the notebook would have done, just without the visualisation. 

Since a deprecated installation of pandas was installed due to python versioning differences, each time a new line of combined codes and processed text was added, a new blank line was also added that made the program throw errors. This was sorted by running the clean_notes program that removed all blank lines. 

The model was then used by running the runTest.py file with the existing code blocks already set up for MIMIC-III.

\section{Full Evaluation Results}
\label{appendix_full_eval}
\begin{figure*}[!t]
\begin{center}
\centering
\includegraphics*[width=0.99\textwidth]{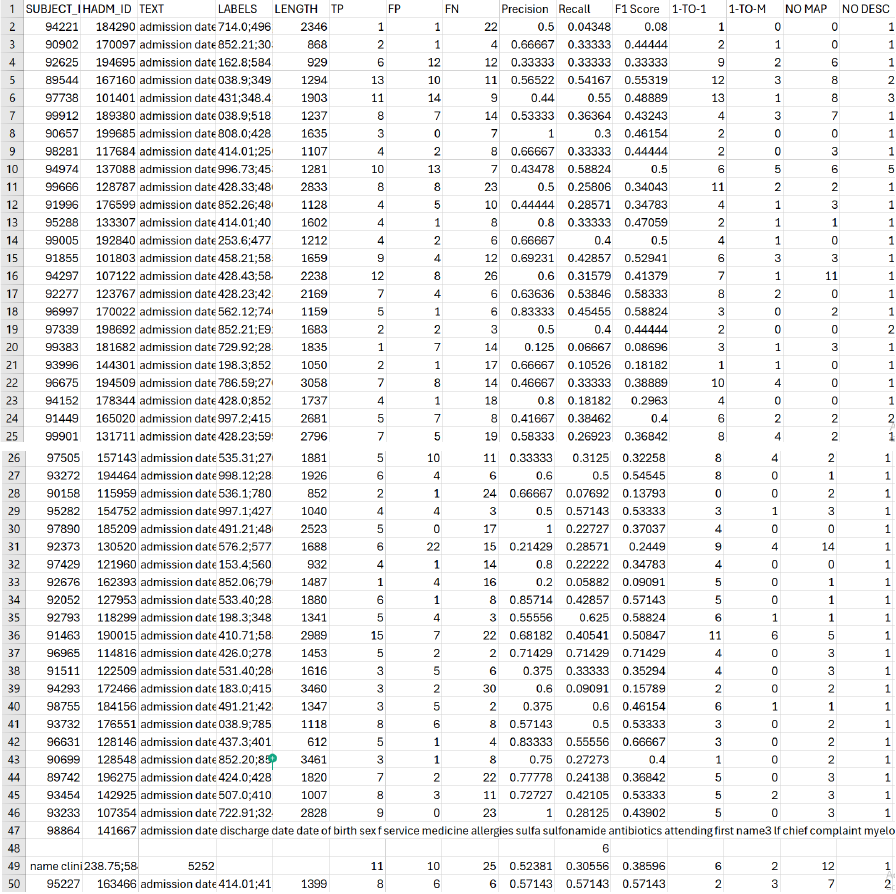}
\caption{Full Evaluation Results - Part 1
}
\label{fig:full-eval-p1}
\end{center}
\end{figure*}

\begin{figure*}[!t]
\begin{center}
\centering
\includegraphics*[width=0.99\textwidth]{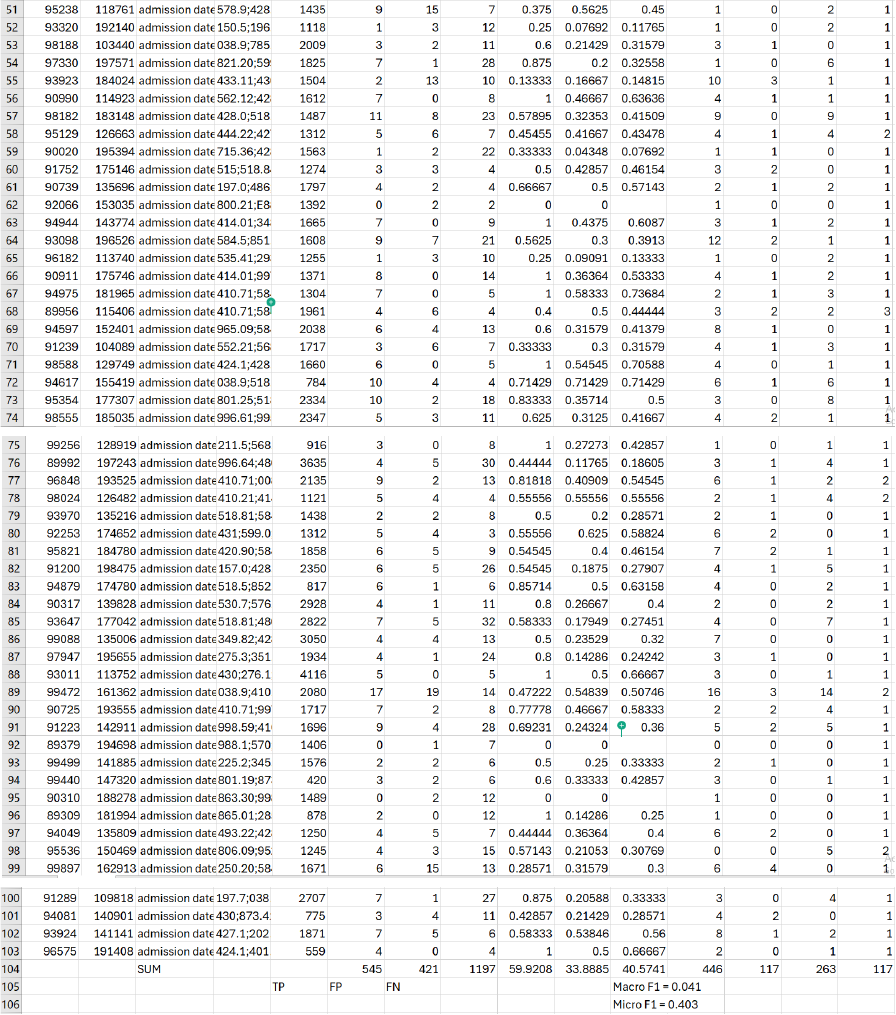}
\caption{Full Evaluation Results - Part 2
}
\label{fig:full-eval-p2}
\end{center}
\end{figure*}

The full evaluation results are listed in Figure \ref{fig:full-eval-p1} and \ref{fig:full-eval-p2}.

\section{Example Letters from Stakeholder and Results}
\label{appendix_letters}

Letter 1:
`` Dear xx xxx,

I saw xxx today in clinic. I think he has chronic otitis media. I have inserted some grommets, which should hopefully improve his hearing.

Yours Sincerely,

xx xxx ''

\begin{figure*}[!t]
\begin{center}
\centering
\includegraphics*[width=0.99\textwidth]{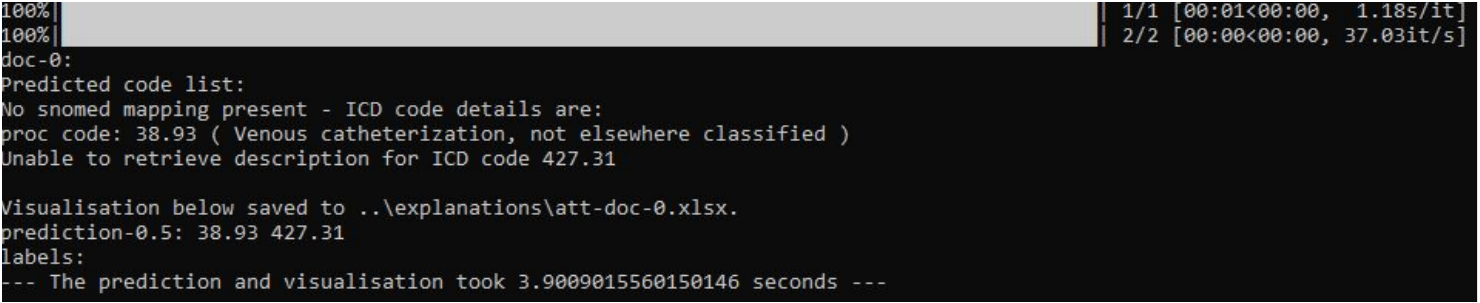}
\caption{Letter1 Outcomes
}
\label{fig:Letter1-outcome-cropped}
\end{center}
\end{figure*}

$\Rightarrow$ Letter 1 (anonymized) result is shown in Figure \ref{fig:Letter1-outcome-cropped}. 
The prediction results for ICD code is ‘proc code 38.93’ (Venous catheterization), prediction 427.31 = atrial fibrillation.

\begin{figure*}[!t]
\begin{center}
\centering
\includegraphics*[width=0.99\textwidth]{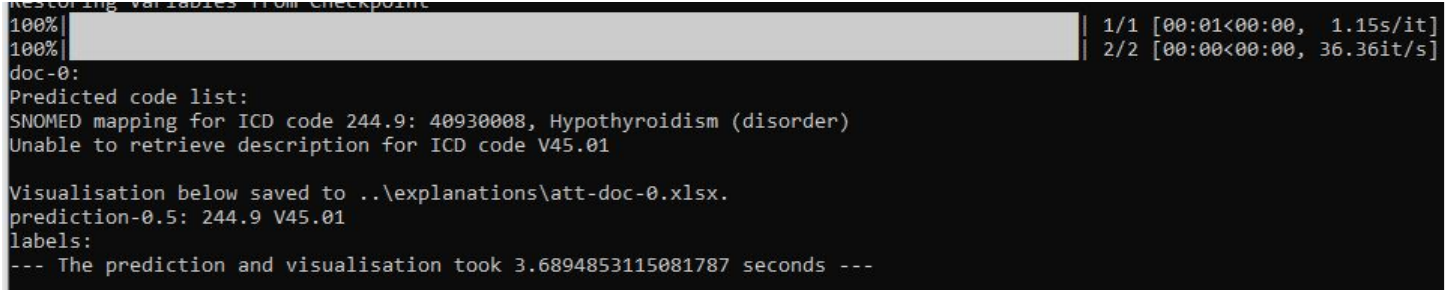}
\caption{Letter2 Outcomes
}
\label{fig:Letter2-outcome-cropped}
\end{center}
\end{figure*}

\begin{figure*}[!t]
\begin{center}
\centering
\includegraphics*[width=0.99\textwidth]{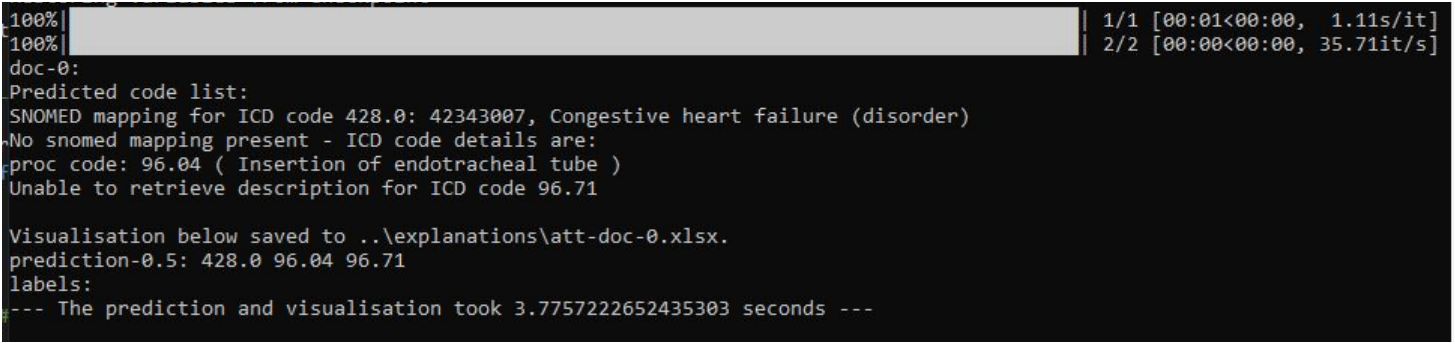}
\caption{Letter3 Outcomes
}
\label{fig:Letter3-outcome-cropped}
\end{center}
\end{figure*}

Letter 2: 
`` Dear xx xxx,

Thank you for sending xxx to me. I agree that I think she has quite bad psoriasis; I will refer her for phototherapy.

Yours Sincerely,

xx xxx xxx ''

$\Rightarrow$ Letter 2 (anonymized) result is shown in Figure \ref{fig:Letter2-outcome-cropped}. 
The prediction result SNOMED mapping for ICD CODE 244.9 \footnote{\url{https://www.findacode.com/icd-9/244-9-hypothyroidism-primary-nos-icd-9-code.html}} is 40930008, which is Hypothyroidism (disorder) \footnote{\url{https://www.findacode.com/snomed/40930008--hypothyroidism.html}}. ICD code V45.01 is cardiac pacemaker in situ \footnote{\url{https://www.findacode.com/icd-9/v45-01-postsurgical-state-cardiac-pacemaker-icd-9-code.html}}. 

Letter 3:
`` Dear xx xxx,

I reviewed xxx following his PCA - this has indeed shown a MI which is clearly causing LVF, as evidenced by his raised BNP. We will proceed to a CABG

xx xxx xxx ''

$\Rightarrow$ Letter 3 (anonymized) result is shown in Figure \ref{fig:Letter3-outcome-cropped}. 
It predicted SNOMED mapping 42343007, which is congestive heart failure (disorder) \footnote{\url{https://bioportal.bioontology.org/ontologies/SNOMEDCT?p=classes&conceptid=42343007}}.
ICD code 96.71 is ``continuous invasive mechanical ventilation for less than 96 consecutive hours'' \footnote{\url{https://www.findacode.com/icd-9/96-71-continuous-mechanical-ventilation-less-than-96-icd-9-procedure-code.html}}.

Dear xxx xxx,

I saw xxx today, he has clearly developed Waldenstroms Macroglubulinaemia, which is unusual given his Tay-Sach's disease. I will start him on chemotherapy shortly.

Best Wishes,

xxx xxx xxx xxx

$\Rightarrow$
No codes found. 

\end{document}